\documentclass[11pt]{article}
\usepackage{colacl}
\usepackage{epsfig}
\title{Unlimited Vocabulary Grapheme to Phoneme Conversion for Korean TTS}
\author{Byeongchang Kim \and WonIl Lee \and Geunbae Lee \and Jong-Hyeok Lee\\
Department of Computer Science \& Engineering\\
Pohang University of Science \& Technology\\
Pohang, Korea\\
\{bckim, bdragon, gblee, jhlee\}@postech.ac.kr }
\begin{document}
\maketitle
\begin{abstract}
	This paper describes a grapheme-to-phoneme conversion method using
	phoneme connectivity and CCV conversion rules. The method consists
	of mainly four modules including morpheme normalization, phrase-break detection,
	morpheme to phoneme conversion and phoneme connectivity check.\\
	The morpheme normalization is to replace non-Korean symbols into standard Korean graphemes.
	The phrase-break detector assigns phrase breaks using part-of-speech (POS) information.
	In the morpheme-to-phoneme conversion module, each morpheme in the phrase is converted into
	phonetic patterns by looking up the morpheme phonetic pattern dictionary which
	contains candidate phonological changes in boundaries of the morphemes.
	Graphemes within a morpheme are grouped into CCV patterns and converted into phonemes by the
	CCV conversion rules. The phoneme connectivity table supports grammaticality checking
	of the adjacent two phonetic morphemes.\\
	In the experiments with a corpus of 4,973 sentences, we achieved 99.9\% of the grapheme-to-phoneme conversion performance and 97.5\% of the sentence conversion
	performance. The full Korean TTS system is now being implemented using this conversion method.\\
\end{abstract}
\section{Introduction}
During the past few years, remarkable improvements have been made for
high-quality text-to-speech systems \cite{santen:progress}. One of the enduring problems in developing
high-quality text-to-speech system is accurate grapheme-to-phoneme conversion \cite{divay:algorithms}.
It can be described as a function mapping the spelling of words to their phonetic
symbols. Nevertheless, the function in some alphabetic languages needs some linguistic knowledge,
especially morphological and phonological, but often also semantic
knowledge.\\
In this paper, we present a new grapheme-to-phoneme conversion method for
unlimited vocabulary Korean TTS. The conversion method is divided into
mainly four modules.  Each module has its own linguistic knowledge.
Phrase-break detection module assigns phrase breaks onto part-of-speech
sequences using morphological knowledge. Word-boundaries before and after
phrase breaks should not be co-articulated. So, accurate phrase-break assignments are essential in high quality TTS systems. In the morpheme-to-phoneme conversion module,
boundary graphemes of each morpheme in the phrase are converted to phonemes
by applying phonetic patterns which contain possible phonological changes in
the boundaries of morphemes. The patterns are designed using morphological and
phonotactic knowledge. Graphemes within a morpheme are converted into
phonemes by CCV (consonant consonant vowel) conversion rules which are automatically extracted from a corpus.
After all the conversions, phoneme connectivity table supports the grammaticality of the
adjacency of two phonetic morphemes. This grammaticality comes from Korean
phonology rules.\\
This paper is organized as follows. Section 2 briefly explains the characteristics
of spoken Korean for general readers. Section 3 and 4 introduces our grapheme-to-phoneme conversion method based on morphological and phonological knowledge of
Korean. Section 5 shows experiment results to demonstrate the performance and
Section 6 draws some conclusions.

\section{Features of Spoken Korean}
This section briefly explains the linguistic characteristics of spoken Korean
before describing the architecture.\\
A Korean word (called eojeol) consists of more than one morpheme
with clear-cut morpheme boundaries (Korean is an agglutinative language).
Korean is a postpositional language with many kinds of noun-endings,
verb-endings, and prefinal verb-endings. These functional morphemes determine
the noun's case roles, verb's aspect/tenses, modals, and modification relations
between words.
The unit of pause in speech (phrase break) is usually different from that in written
text. No phonological change occur between these phrase breaks.
Phonological changes can occur in a morpheme, between morphemes in a word,
and even between words in a phrase break as described in the 30 general phonological rules for Korean\cite{korean:rule}. These changes include consonant and
vowel assimilation, dissimilation, insertion, deletion, and contraction.
For example, noun ``kag-ryo'' pronounced as ``ka\textbf{ngn}yo'' (meaning ``cabinet'')
is an example of phonological change within a morpheme. Noun plus noun-ending ``such+gwa'', in which ``such'' means
``charcoal'' and ``gwa'' means ``and'' in English, is sounded as ``su\textbf{dgg}wa'', which is an example of the
inter-morpheme phonological change. ``Ta-seos gae'', which means
``five items'', is sounded as ``taseo\textbf{t gg}ae'', in which phonological changes
occur between words.
In addition, phonological changes can occur conditionally on the
morphotactic environments but also on phonotactic environments.

\section{Architecture of the Grapheme-to-Phoneme Converter}
\begin{figure}[th]
\frame{\centerline{\epsfig{file=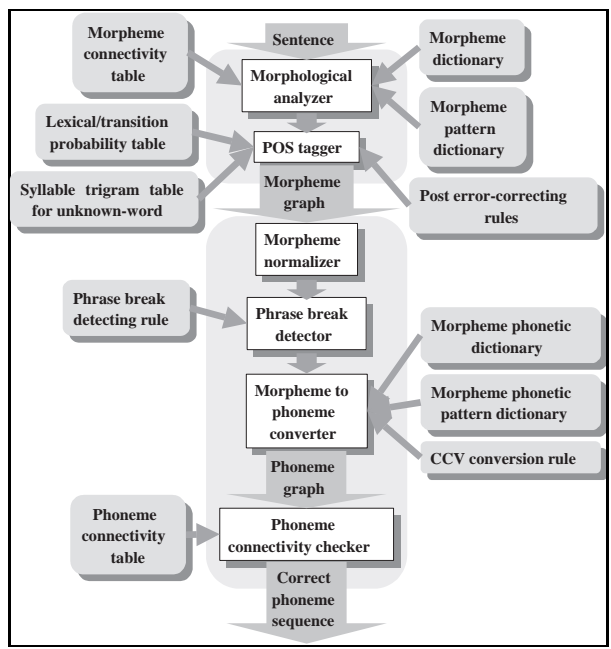,width=8cm}}}
\caption{Architecture of the grapheme-to-phoneme converter in TTS applications}
\label{system}
\end{figure}

Part-of-speech (POS) tagging is a basic step to the grapheme-to-phoneme conversion since
phonological changes depend on morphotactic and phonotactic environments.
The POS tagging system have to handle out-of-vocabulary (OOV) words for accurate
grapheme-to-phoneme conversion of unlimited vocabulary \cite{bechet:automatic}. Figure \ref{system} shows the
architecture of our grapheme-to-phoneme converter integrated with the hybrid POS
tagging system \cite{lee:hybridP}. The hybrid POS tagging system employs generalized OOV word handling
mechanisms in the morphological analysis, and cascades statistical and rule-based approaches
in the two-phase training architecture for POS disambiguation.\\

Each morpheme tagged by the POS tagger is normalized by replacing non-Korean symbols by Korean
graphemes to expand numbers, abbreviations, and acronyms. The phrase-break
detector segments the POS sequences into several phrases according to phrase-break detection rules. In the phoneme converter, each morpheme in the phrase is converted into phoneme
sequences by consulting the morpheme phonetic dictionary. 
The OOV morphemes which are not registered in the morpheme phonetic dictionary should be processed in two different ways. The graphemes in the morpheme boundary
are converted into phonemes by consulting the morpheme phonetic pattern dictionary.
The graphemes within morphemes are converted into phonemes according to CCV conversion rules. To model 
phoneme's connectablities between morpheme boundaries, the separate phoneme connectivity table
encodes the phonological changes between the morpheme with their POS tags. Outputs of
the grapheme-to-phoneme converter, that is, phoneme sequences of the input sentence,
can be directly fed to the lower level signal processing module of TTS systems.
Next section will give detail descriptions of each component of the grapheme-to-phoneme
converter. The hybrid POS tagging system will not be explained in this paper,
and interested readers can see the reference \cite{lee:hybridP}.

\section{Component Descriptions of the Converter}
\subsection{Morpheme Normalization}
The normalization replaces non-Korean symbols by corresponding Korean graphemes.
Non-Korean symbols include numbers (e.g. 54, -12, 5,400, 4.2), dates (e.g. 20/1/97, 20-Jan-97),
times (e.g. 12:46), scores (e.g. 74:64), mathematical expressions (e.g. 4+5, 1/3), telephone numbers,
abbreviations (e.g. km, ha) and acronyms (e.g. UNESCO, OECD). Especially, acronyms
have two types: spelled acronyms such as OECD and pronounced ones like a word
such as UNESCO.\\
The numbers are converted into the corresponding Korean graphemes using deterministic
finite automata. The dates, times, scores, expressions and telephone numbers
are converted into equivalent graphemes using their formats and values. The
abbreviations and acronyms are enrolled in the morpheme phonetic dictionary, and
converted into the phonemes using the morpheme-to-phoneme conversion module.

\subsection{Phrase-Break Detection}
Phrase-break boundaries are important to the subsequent processing such as morpheme-to-phoneme conversion and prosodic feature generation. Graphemes in
phrase-break boundaries are not phonologically changed and sounded as their original
corresponding phonemes in Korean. \\
A number of different algorithms have been suggested and implemented for phrase
break detection \cite{black:assigning}. The simplest algorithm uses deterministic rules and more
complicated algorithms can use syntactic knowledge and even semantic knowledge.
We designed simple rules using break and POS tagged corpus. We found that, in Korean,
the average
length of phrases is 5.6 words and over 90\% of breaks are after 6 different POS
tags: conjunctive ending, auxiliary particle, case particle, other
particle, adverb and adnominal ending. The phrase-break detector assigns
breaks after these 6 POS tags considering the length of phrases.\\

\subsection{Morpheme-to-Phoneme Conversion}
\begin{figure}[th]
\frame{\centerline{\epsfig{file=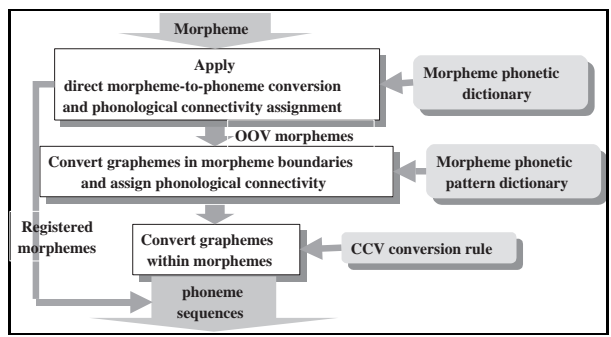,width=8cm}}}
\caption{Morpheme-to-phoneme conversion for unlimited vocabularies}
\label{converter}
\end{figure}

The morphemes registered in the morpheme phonetic dictionary can be directly
converted into phonemes by consulting the dictionary entries. However,
separate method to process the OOV morphemes which are not registered in the
dictionary is necessary. We developed a new method as shown Figure \ref{converter}.

\begin{table*}[th]
\caption{Example entries of the morpheme phonetic dictionary}
\label{morpheme-phonetic-dictionary-example}
\begin{center}
\begin{tabular}{|c|c|c|c|c|} \hline
POS tag & morpheme & phoneme sequence & left connectivity & right connectivity \\ \hline
common noun & pang-gabs & pang-ggam & `p' no change & `bs' changed to `m' \\ \hline
common noun & pang-gabs & pang-ggab & `p' no change & `bs' changed to `b' \\ \hline
common noun & pang-gabs & pang-ggabss & `p' no change & `bs' changed to `bss' \\ \hline
\end{tabular}
\end{center}
\end{table*}

\begin{table*}[th]
\caption{Example entries of morpheme phonetic pattern dictionary}
\label{morpheme-phonetic-pattern-dictionary-example}
\begin{center}
\begin{tabular}{|c|c|c|c|c|} \hline
POS tag & morpheme & phoneme sequence & left connectivity & right connectivity \\ \hline
irregular verb & t$\ast$d & tt$\ast$n & `t' changed to `tt' & `d' changed to `n' \\ \hline
irregular verb & t$\ast$Z & tt$\ast$Z & `t' changed to `tt' & no change \\ \hline
irregular verb & Y$\ast$d & Y$\ast$n & no change & `d' changed to `n' \\ \hline
irregular verb & Y$\ast$Z & Y$\ast$Z & no change & no change \\ \hline
\end{tabular}
\end{center}
\end{table*}

The morpheme phonetic dictionary contains POS tag, morpheme, phoneme
connectivity (left and right) and phoneme sequence for each entry. We try to register
minimum number of morpheme in the dictionary. So it contains only the
morphemes which are difficult to process using the next OOV morpheme conversion modules.
Table \ref{morpheme-phonetic-dictionary-example} shows example entries for the
common noun ``pang-gabs'', meaning ``price of a room'' in hotel reservation dialogs.
The common noun ``pang-gabs'' can be
pronounced as ``pang-ggam'', ``pang-ggab'' or ``pang-ggabss''
according to first phoneme of the adjacent morphemes.\\
To handle the OOV morphemes, morpheme
phonetic {\em pattern} dictionary is developed to contain all the general patterns of Korean POS tags,
morphemes, phoneme connectivity and phoneme sequences. Boundary phonemes of
the OOV morphemes can be converted to their candidate phonemes, and the phonological
connectivity for them can be acquired by consulting this morpheme phonetic
pattern dictionary. Example entries corresponding to the irregular verb ``teud'',
meaning ``hear'', are shown in
Table \ref{morpheme-phonetic-pattern-dictionary-example}. Meta characters,
`Z', `Y', `V', `$\ast$' designate single consonant, consonant except silence
phoneme, vowel, any character sequence with variable length in the order. The table shows
that the first grapheme `t' can be phonologically changed to `tt'
according to the last phoneme of the preceding morpheme (left connectivity), and the last
grapheme `d' can be phonologically changed to `n' according to
the first phoneme of the following morpheme(right connectivity). The morpheme phonetic pattern
dictionary contains similar 1,992 entries to model the general phonological rules for Korean.\\
The graphemes within a morpheme for OOV morphemes are converted into phonemes using the CCV conversion rules.
The CCV conversion rules are the mapping rules between grapheme to phoneme in character tri-gram forms
which are in the order of consonant(C) consonant(C) vowel(V) spanning two 
consecutive syllables. The CCV rules are designed and automatically learned 
from a corpus reflecting the following Korean phonological facts.

\begin{itemize}
\item Korean is a syllable-base language, i.e., Korean syllable is the basic
unit of the graphemes and consists of first consonant, vowel and final consonant (CVC).
\item The number of possible consonants for each syllable can be varied in grapheme-to-phoneme conversion.
\item The number of vowels for each syllable is not changed. 
\item Phonological changes of the first consonant are only affected by the final
consonant of the preceding syllable and the following vowel of the same syllable.
\item Phonological changes of the final consonant are only affected by the first
consonant of the following syllable.
\item Phonological changes of the vowel are not affected by the following
consonant.
\end{itemize}

The boundary graphemes of the OOV morphemes are phonologically changed according to the
POS tag and the boundary graphemes of the preceding and following morphemes. On the
other hand, the inner grapheme conversion is not affected by the POS tag,
but only by the adjacent graphemes within the same morpheme. The CCV conversion rules
can model the fact easily, but the conventional CC conversion rules \cite{park:implementation}
cannot model the influence of the vowels.

\subsection{Phoneme Connectivity Check}
To verify the boundary phonemes' connectablity to one another, the separate
phoneme connectivity table encodes the phonologically connectable pair of each morpheme which has
phonologically changed boundary graphemes. This phoneme connectivity
table indicates the grammatical sound combinations in Korean phonology using the
defined left and right connectivity information.\\
The morpheme-to-phoneme conversion can generate a lot of phoneme sequence candidates for single morpheme.
We put the whole phoneme sequence candidates in a phoneme graph where a correct phoneme
sequence path can be selected for input sentence. The phoneme connectivity
check performs this selection and prunes the ungrammatical phoneme sequences in the graph.

\section{Implementation and Experiment Results}
We implemented simple phrase-break detection rules from break and POS tagged
corpus collected from recording and transcribing broadcasting news. The rules reflect the fact that average
length of phrases in Korean is 5.6 words and over 90\% of breaks are after 6 specific POS
tags, described in the texts.\\
We constructed a 1,992 entry morpheme phonetic pattern dictionary for
OOV morpheme processing using standard Korean phonological rules. The morpheme phonetic
dictionary was constructed for only the morphemes that are difficult to handle with
these standard rules. The two
dictionaries are indexed using POS tag and morpheme pattern for fast access. To model the
boundary phonemes' connectablity to one another, the phoneme connectivity
table encodes 626 pair of phonologically connectable morphemes.\\
The 2030 entry rule set for CCV conversion was automatically learned from
phonetically transcribed 9,773 sentences. The independent phonetically transcribed
4,973 sentences are used to test the performance of the grapheme-to-phoneme
conversion. Of the 4,973 sentences, only 2.5\% are incorrectly processed (120 sentences out of
4,973), and only 0.1\% of the graphemes in the sentences are actually incorrectly converted.

\section{Conclusions}
This paper presents a new grapheme-to-phoneme conversion method using phoneme
connectivity and CCV conversion rules for unlimited vocabulary Korean TTS.
For the efficient conversion, new ideas of morpheme phonetic and morpheme phonetic
pattern dictionary are invented and the system demonstrates remarkable conversion
performance for the unlimited vocabulary texts.
Our main contributions include presenting the morphologically and phonologically
conditioned conversion model which is essential for morphologically
and phonologically complex agglutinative languages.
The other contribution is the grapheme-to-phoneme conversion model
combined with the declarative phonological rule which is well suited to the
given task. We also designed new CCV unit of
grapheme-to-phoneme conversion for unlimited vocabulary task. The experiments show
that grapheme-to-phoneme conversion performance is 97.5\% in
sentence conversion, and 99.9\% in each grapheme conversion. We are
now working on incorporating this grapheme-to-phoneme conversion into the developing TTS systems.
\bibliographystyle{acl}

\end{document}